\documentclass{article}

\usepackage{arxiv}

\usepackage[utf8]{inputenc} 
\usepackage[T1]{fontenc}    
\usepackage{hyperref}       
\usepackage{url}            
\usepackage{booktabs}       
\usepackage{amsfonts}       
\usepackage{nicefrac}       
\usepackage{microtype}      
\usepackage{lipsum}
\usepackage{graphicx}
\graphicspath{ {./images/} }

\title{Comparative Analysis of AutoML and BiLSTM Models for Cyberbullying Detection on Indonesian Instagram Comments}

\author{
Raihana Adelia Putri \\
Institut Teknologi Sumatera\\
  \texttt{raihana.123450041@student.itera.ac.id} \\
   \And
 Aisyah Musfirah \\
  Institut Teknologi Sumatera\\
  \texttt{aisyah.123450084@student.itera.ac.id} \\
  \And
 Anggi Puspita Ningrum \\
  Institut Teknologi Sumatera\\
  \texttt{anggi.123450012@student.itera.ac.id} \\
  \And
   \And
  Luluk Muthoharoh, M.Si \\
  Institut Teknologi Sumatera\\
  \texttt{luluk.muthoharoh@sd.itera.ac.id} \\
  \And
  Ardika Satria, M.Si \\
  Institut Teknologi Sumatera\\
  \texttt{ardika.satria@sd.itera.ac.id} \\
  \And
  Martin Clinton Tosima Manullang, Ph.D. \\
  Institut Teknologi Sumatera\\
  \texttt{martin.manullang@if.itera.ac.id} \\
}
\usepackage{graphicx}
\usepackage{float}
\usepackage{hyperref}

\begin{document}
\maketitle
\begin{abstract}
Cyberbullying on social media platforms, particularly Instagram, poses a serious threat to mental health, especially among Indonesian youth. This study presents a comparative analysis of machine learning (ML) and deep learning (DL) models for detecting cyberbullying in Indonesian-language Instagram comments. Three ML algorithms Naive Bayes, Logistic Regression, and Support Vector Machine (SVM) using TF-IDF feature representation are compared against two DL architectures Bi-LSTM and Bi-LSTM with Bahdanau Attention mechanism. The dataset consists of 650 balanced and binary-labeled comments (Bullying and Non-Bullying) sourced from the accounts of Indonesian artists and influencers. A preprocessing pipeline tailored for informal Indonesian text was applied, incorporating slang normalization, stopword removal, and morphological stemming using Indonesian-specific NLP libraries. Experimental results show that Logistic Regression achieved the best performance among ML models with an accuracy of 85.25\% and F1-score of 85.22\%, while Bi-LSTM with Attention outperformed all models with an accuracy of 84.62\% and F1-Macro of 84.58\%. Although DL models demonstrated superior contextual understanding, ML models remain competitive for resource-constrained deployments. These findings underscore the importance of domain-specific preprocessing and provide a practical foundation for developing automated content moderation systems on Indonesian-language social media platforms.
 
\end{abstract}

\keywords{Cyberbullying Detection \and Instagram \and Machine Learning \and Deep Learning \and Bi-LSTM \and TF-IDF \and Indonesian NLP \and Text Classification}

\section{Introduction}
Social media, especially Instagram, has become a breeding ground for cyberbullying the intentional and repeated use of digital technology to harass, threaten, or humiliate someone [1]. In the second quarter of 2024, Instagram recorded more than 10.1 million instances of bullying and harassment worldwide [2]. Indonesia is taking this issue seriously, as 45\% of the 2,777 young people aged 14-24 who were surveyed in 2022 reported having experienced cyberbullying [3], with consequences ranging from anxiety and depression to cases of suicide among adolescents.

Many automated approaches have been developed to detect cyberbullying using NLP and machine learning. Algorithms such as Naive Bayes, Logistic Regression, and SVM with TF-IDF feature representation have proven effective in text classification [4][5], meanwhile, the Bi-LSTM deep learning model is capable of capturing more complex patterns by processing text sequences in both directions simultaneously [6]. Research on Indonesian social media data shows an accuracy of over 88\% using a combined CNN and Bi-LSTM model [7]. However, most studies still focus on Twitter/X data, while Instagram comments are rich in slang, emojis, and mixed languages. This has rarely been studied comparatively between conventional ML and deep learning approaches [8].

The main contributions of this study are: (1) a comprehensive exploration and comparison of three ML algorithms (Naive Bayes, Logistic Regression, SVM) and one DL model (Bi-LSTM) on an Indonesian-language cyberbullying dataset; (2) an analysis of the effectiveness of the NLP preprocessing pipeline on model performance; and (3) the identification of the best model that can serve as the foundation for developing an automated content moderation system on Indonesian-language social media platforms.

\section{Related Work}
\label{sec:headings}
Several public datasets are available to support research on the detection of Indonesian-language cyberbullying. One of the first Indonesian-language hate speech datasets was developed by Alfina et al. from Twitter, and has since been widely used as a reference in Indonesian NLP research [9]. The dataset used in this study is “Cyberbullying Bahasa Indonesia,” curated by Rahman et al. [10]. The Indonesian-language text in this dataset has been annotated with binary labels: Bullying and Non-Bullying. The availability of labeled datasets like this is crucial for training and evaluating supervised classification models.

To identify cyberbullying in Indonesian-language text, various machine learning approaches have been employed. Nugraha and Astuti used Support Vector Machines (SVMs) with TF-IDF feature representation on Instagram comments and demonstrated that this technique can be applied to visual platforms such as Instagram [11]. In contrast, Sheth et al. investigated various machine learning algorithms, including Naive Bayes, Logistic Regression, and SVM, for text classification tasks. They found that each algorithm has unique advantages depending on the characteristics of the dataset [5]. These findings reinforce the importance of comparing models to determine the most suitable algorithm, as done in this study.

A number of studies have shown that neural network-based models are better at capturing more complex textual contexts, in line with advances in deep learning. In a systematic review conducted by Yao et al., they found that LSTM-based models and their variants, including Bidirectional LSTM (Bi-LSTM), consistently outperformed conventional machine learning models in cyberbullying detection tasks [6]. Nasution and Setiawan demonstrated that, for the Indonesian language, combining FastText with CNN and Bi-LSTM architectures improves cyberbullying detection on Indonesian Twitter data [12]. Nevertheless, Asqolani and Setiawan showed that using Word2Vec as a feature expansion in a hybrid deep learning model can significantly enhance detection [8]. In addition, Hafizh Fattah et al. confirmed that a combined CNN-LSTM model achieved an accuracy of over 88\% on an Indonesian-language cyberbullying dataset from the X platform [7]. This study expands upon that work by applying Bi-LSTM and directly comparing it with classical machine learning models on Indonesian-language Instagram comment data.

\section{Dataset}
\label{sec:headings}

\subsection{Dataset Description}
The dataset used in this study consists of a collection of Instagram user comments gathered from the accounts of Indonesian artists and influencers, with a focus on identifying instances of cyberbullying. This dataset consists of 650 comment records stored in a CSV format with a semicolon as the delimiter, and includes six main attributes: sequence number, Instagram username of the commenter, comment content, label category, posting date, and the Instagram username of the artist or influencer targeted by the comment.

In terms of labeling, this dataset is balanced with an even distribution between two classes: 325 data points labeled “Bullying” and 325 labeled “Non-bullying,” thus avoiding class imbalance that could affect the performance of the classification model. 

In terms of text characteristics, the comments are written in informal Indonesian with an average length of 63 characters per comment. Comment length varies from 24 to 173 characters, with a median of 55 characters and a standard deviation of 28.41. This indicates that most comments are brief and concise, as is typical of social media interactions. This dataset is suitable for use as a training and evaluation basis for Natural Language Processing (NLP) models for binary text classification tasks, particularly in the context of detecting Indonesian-language cyberbullying on the Instagram platform.

\subsection{Statistic}
The dataset consists of 650 comments with a balanced label distribution between the Bullying and Non-Bullying classes (325 data points each, or 50\%). Comment length varies, with a right-skewed distribution typical of social media data, where comments labeled as Bullying tend to be more lexically dense and negatively charged compared to Non-Bullying comments, which are more varied [6]. The dataset includes comments from various accounts of Indonesian artists and social media influencers with an uneven distribution across accounts, providing linguistic contextual diversity that supports the model’s generalization ability. Word frequency analysis reveals clear lexical differences between the two classes. The Bullying class is dominated by words containing personal attacks, while the Non-Bullying class is dominated by expressions of praise and support. This reinforces the potential of a TF-IDF-based approach. Data quality checks confirmed no duplicates or missing values, so the dataset is deemed clean and ready for use without additional imputation steps.

\begin{table}[H]
\caption{Average Number of Words per Comment by Category}
\centering
\begin{tabular}{lc}
\toprule
Category & Average Number of Words \\
\midrule
Bullying & 9.48 \\
Non-bullying & 11.34 \\
\bottomrule
\end{tabular}
\label{tab:rata_kata}
\end{table}

\subsection{Pre-Processing}
The preprocessing stage is a crucial step in the NLP pipeline to ensure the quality of the model’s input, given that Indonesian-language Instagram comments have distinctive characteristics such as slang, informal abbreviations, excessive character repetition, and a mix of emojis that can significantly degrade model performance without adequate handling [8]. The preprocessing pipeline was developed using three primary Indonesian-language libraries: indoNLP for slang normalization and character elongation, nlp-id for stopword removal, and PySastrawi for Indonesian morphological stemming, which were selected for their ability to handle linguistic complexities that generic NLP tools cannot address [14].

\begin{table}[H]
\caption{Text Preprocessing Pipeline Stages}
\centering
\begin{tabular}{clll}
\toprule
No & Stage & Description & Library/Function \\
\midrule
1 & Case Folding & Converts all text to lowercase letters & text.lower() \\
2 & Cleaning & Removes mentions, hashtags, URLs, and non-alphabetic characters & re.sub(...) \\
3 & Slang Normalization & Replaces slang words and character elongations (e.g., bgt $\rightarrow$ banget) & replace\_slang(), replace\_word\_elongation() \\
4 & Stopword Removal & Removes common Indonesian words that are not discriminative & StopWord.remove\_stopword() \\
5 & Stemming & Reduces words to their root forms based on Indonesian morphology & stemmer.stem() \\
6 & Tokenization & Splits cleaned text into a list of tokens & str.split() \\
\bottomrule
\end{tabular}
\label{tab:preprocessing_pipeline}
\end{table}

\section{Methodology}

The research stages begin with the acquisition of Instagram comment data, which undergoes an intensive text preprocessing phase to handle non-standard language characteristics, such as the use of numbers as letters (\textit{leetspeak}) and slang expansion (\textit{slang}). This process is crucial to ensure the model can accurately capture the semantic context of bullying before entering the parallel modeling stage.

\begin{figure}[H]
    \centering
    \includegraphics[width=0.45\textwidth]{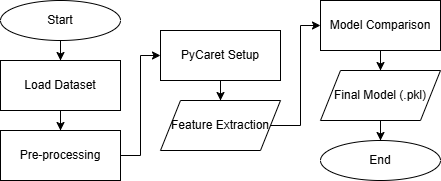}
    \caption{Systematic Stages of the \textit{Machine Learning} Path.}
    \label{fig:flowchart_ml}
\end{figure}

The first workflow, the machine learning path as illustrated in Figure \ref{fig:flowchart_ml}, starts with the Load Dataset stage followed by Preprocessing for thorough text cleaning. The next step is the PyCaret Setup to define features and targets. The process then proceeds to the Model Comparison stage to evaluate various baseline algorithms simultaneously, followed by Hyperparameter Tuning on the selected model. Once optimal performance is achieved, the model is finalized through the finalize model stage and re-evaluated before being ready for the Deployment stage in (.pkl) file format.

\begin{figure}[H]
    \centering
    \includegraphics[width=0.50\textwidth]{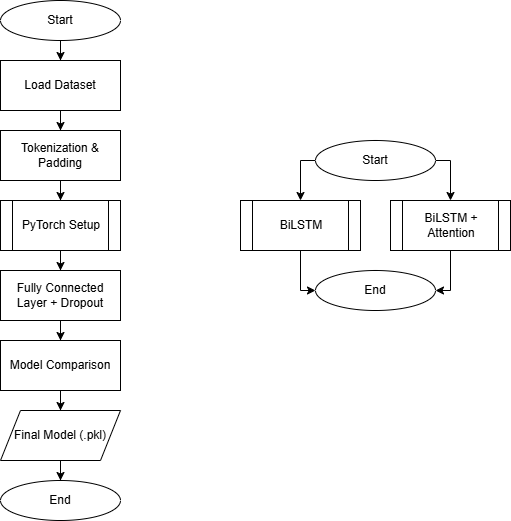}
    \caption{Stages of the \textit{Deep Learning} Architecture.}
    \label{fig:flowchart_dl}
\end{figure}

Meanwhile, the deep learning path in Figure \ref{fig:flowchart_dl} follows more complex stages to capture sequential relationships. After passing through text preprocessing, the data enters the Tokenization \& Padding stage to standardize sequence lengths. The process is then divided into two main architectural paths: Model 1 uses a BiLSTM Classifier for bidirectional context, Model 2 adds a Bahdanau Attention layer for keyword weighting. All these models undergo the Model Training \& Evaluation stage with a k-fold cross validation scheme to determine the most optimal model for predicting bullying and non-bullying labels.

\section{Experiments}

\subsection{Computing Environment}
All experiments were executed on a system based on a Central Processing Unit (CPU) using the Python programming language $\leq$ 3.11. The primary libraries used include PyCaret for automated machine learning (Auto ML), as well as PyTorch and Transformers for deep learning model development.

\subsection{Hyperparameter Settings}
Parameter settings were conducted through two different approaches according to the chosen modeling path. In the machine learning path, hyperparameter tuning was performed automatically via the PyCaret framework using the Grid Search method, focused on maximizing the F1-Score with TF-IDF based feature extraction. Meanwhile, for the deep learning path, the configuration was specifically determined as detailed in Table \ref{table:dl_config}.

\begin{table}[H]
\caption{Hyperparameter Configuration for Deep Learning Models}
\label{table:dl_config}
\centering
\begin{tabular}{lc}
\toprule
Parameter & Value \\
\midrule
Batch Size & 32 \\
Embedding Size & 128 \\
Learning Rate & 0.001 \\ 
Optimizer & Adam \\ 
Loss Function & Cross Entropy Loss \\ 
Maximum Epochs & 15 \\ 
Patience (Early Stopping) & 3 \\ 
Vocabulary Size & 1,368 unique words \\ 
\bottomrule
\end{tabular}
\label{tab:rata_kata}
\end{table}

The configuration above indicates that a batch size of 32 and an embedding size of 128 were chosen to balance computational efficiency and the richness of semantic representation. The use of a 0.001 learning rate with the Adam optimizer ensures stable weight updates. Furthermore, an early stopping mechanism with a patience of 3 was implemented to prevent overfitting by terminating the process if no significant improvement was observed in the validation loss.

\subsection{Validation Strategy and Data Partitioning}
To ensure model stability, different validation strategies were used for the two modeling paths. The machine learning path implemented a 5-Fold Cross Validation scheme on the training data to ensure the model possesses good generalization capabilities. On the other hand, the deep learning path, specifically for the BiLSTM and BiLSTM + Attention architectures, used static data partitioning with a ratio of 80\% training, 10\% validation, and 10\% testing. The entire process was locked using random seed 42 to maintain consistency and reproducibility.

\subsection{Evaluation Metrics}
Model success was measured using several key statistical metrics to provide a comprehensive performance overview. The Accuracy metric was used to determine the ratio of correct predictions, while Precision and Recall monitored the model's ability to minimize detection errors and capture all bullying cases accurately. As the primary indicator, the F1-Score was utilized to provide a balanced assessment of the model's capability in handling cyberbullying text classification, especially in balancing the trade-off between Precision and Recall.

\section{Results \& Discussion}

This section presents the experimental results of the various models used, namely three Machine Learning algorithms (Naive Bayes, Logistic Regression, and Support Vector Machine) and Deep Learning models (Bi-LSTM and its variants). Evaluation was conducted using accuracy, precision, recall, and F1-score metrics with a 10-fold cross-validation scheme to ensure robust results.

\subsection{Experimental Results \& Analysis}
Based on the training and testing results, the TF-IDF-based Machine Learning models demonstrated fairly good performance in classifying comments. As shown in Table \ref{tab:benchmark_ml}, Logistic Regression emerged as the best performing traditional model, achieving the highest scores across all metrics with an accuracy of 0.8525 and an F1-score of 0.8522. Support Vector Machine (SVM) followed closely with an accuracy of 0.8261 and a precision of 0.8445, indicating a strong ability to minimize false positives. In contrast, Naive Bayes showed the lowest performance among the three, with an accuracy of 0.7799, though it remained computationally efficient.

\begin{table}[H]
\centering
\caption{Machine Learning Model Performance Comparison}
\begin{tabular}{lcccc}
\toprule
Model & Accuracy & Precision & Recall & F1-Score \\
\midrule
Naive Bayes         & 0.7799 & 0.7839 & 0.7799 & 0.7789 \\
Logistic Regression & 0.8525 & 0.8553 & 0.8525 & 0.8522 \\
SVM                 & 0.8261 & 0.8445 & 0.8261 & 0.8222 \\
\bottomrule
\end{tabular}
\label{tab:benchmark_ml}
\end{table}

\begin{table}[H]
\centering
\caption{Deep Learning Model Performance Comparison}
\begin{tabular}{lcccc}
\toprule
Model & Accuracy & F1 Macro & F1 Weighted \\
\midrule
BiLSTM                & 0.7846 & 0.7821 & 0.7817 \\
BiLSTM+Attention      & 0.8462 & 0.8458 & 0.8457 \\
\bottomrule
\end{tabular}
\label{tab:benchmark_dl}
\end{table}

On the other hand, Deep Learning models demonstrated their capability in capturing word sequence context more effectively. According to Table \ref{tab:benchmark_dl}, the standard BiLSTM model achieved an accuracy of 0.7846. However, the integration of the attention mechanism in the BiLSTM+Attention model significantly improved the results, reaching an accuracy of 0.8462 and an F1 Macro of 0.8458. This improvement of approximately 6\% in accuracy highlights the importance of the attention mechanism in identifying emotionally significant words within a sentence. Furthermore, the high F1 Weighted score of 0.8457 in the BiLSTM + Attention model confirms its stability in handling the classification task compared to the baseline BiLSTM architecture.

The benchmark table indicates that DL models generally outperform Machine Learning models in terms of accuracy and F1-Score. This is due to the DL models' ability to understand complex linguistic contexts, especially in social media data containing slang, abbreviations, and informal language variations. Nevertheless, Machine Learning models have advantages regarding computational efficiency and faster training times. For small datasets like the one used in this study (650 data points), ML performance remains quite competitive and can be a practical choice.

\subsection{Discussion}
The research findings show that preprocessing plays a vital role in enhancing model performance. Slang normalization and stemming were proven to help the models understand text patterns better. Additionally, the use of a balanced dataset contributed to the stability of model performance. Although Deep Learning models provided the best results, their complexity and high computational requirements are considerations for real-world implementation. Therefore, model selection must be tailored to the system's needs, both in terms of accuracy and efficiency.

\section{Conclusion}

Based on the experimental results and analysis, it can be concluded that Deep Learning models, specifically Bi-LSTM with an attention mechanism, are the best models for detecting cyberbullying in Indonesian Instagram comments. This model is capable of capturing complex linguistic contexts and delivers the highest performance compared to other models. However, Machine Learning models such as Logistic Regression and SVM remains a good alternative as they are lightweight and computationally efficient, especially for implementations with limited resources. This study also reaffirms the importance of text preprocessing in improving the quality of classification models. For future research, it is suggested to use larger datasets and explore more complex Transformer-based models to enhance cyberbullying detection performance.

This study has several limitations worth noting. First, the dataset consists of only 650 comments, which may limit the generalizability of the findings, particularly for deep learning models that typically require larger training corpora. Second, the dataset is sourced exclusively from the accounts of Indonesian artists and influencers, which may not fully represent the diversity of cyberbullying patterns across broader Instagram communities. Third, the binary annotation scheme (Bullying vs. Non-Bullying) does not capture the nuanced categories of cyberbullying such as body shaming, hate speech, or identity-based harassment. 

\bibliographystyle{unsrt}  


\end{document}